\pdfoutput=1
\documentclass[11pt]{article}

\usepackage[]{acl}
\usepackage{times}
\usepackage{latexsym}

\usepackage[T1]{fontenc}
\usepackage[utf8]{inputenc}

\usepackage{microtype}

\usepackage{inconsolata}
\usepackage{graphicx}
\usepackage{amsmath}
\usepackage{amssymb}
\usepackage{multirow}
\usepackage{makecell}
\usepackage{float}
\usepackage{hyphenat}
\title{Multi-stage Retrieve and Re-rank Model for Automatic Medical Coding Recommendation}

\author{Xindi Wang$^{1,2}$, Robert E. Mercer$^{1}$, Frank Rudzicz$^{2,3,4}$\\
$^1$ Department of Computer Science, University of Western Ontario, Canada\\
$^2$ Vector Institute for Artificial Intelligence, Canada\\
$^3$ Faculty of Computer Science, Dalhousie University, Canada\\
$^4$ Department of Computer Science, University of Toronto, Canada\\
xwang842@uwo.ca, mercer@csd.uwo.ca, frank@dal.ca}

\begin{document}
\maketitle
\begin{abstract}
The International Classification of Diseases (ICD) serves as a definitive medical classification system encompassing a wide range of diseases and conditions. %It is used for both clinical and administrative purposes. 
The primary objective of ICD indexing is to allocate a subset of ICD codes to a medical record, which facilitates standardized documentation and management of various health conditions. Most existing approaches have suffered from selecting the proper label subsets from an extremely large ICD collection %(almost 9,000 in the MIMIC-III dataset) 
with a heavy long-tailed label distribution. %(i.e., many labels occur only a few times in the training set). 
%In this paper, unlike previous work that treated ICD indexing as a multi-label classification task, we propose a retrieve-reduce-rerank framework to formalize the task as a recommendation task, where the objective is to produce a list of recommended ICD codes for a medical record. 
%The three-stage pipeline includes an ICD retrieval step using the auxiliary knowledge (i.e., different clinical code terminologies and drug prescriptions) of electronic health records (EHR) that prepares a small candidate set for scoring. The reduction stage, using BM25, scores the candidates and further shortens the list of relevant ICD codes. %, which initially allows partial lexical matching between clinical texts and ICD codes. 
%In the final step, we propose a label co-occurrence guided contrastive re-ranking model. By pulling together the clinical notes with the corresponding positive ICD codes, the re-ranking model implicitly acquires the code co-occurrence information from the label representations independently and then matches with the representations of the clinical notes, which can extract more separated features with different labels that further improves the recommendation list.
In this paper, we leverage a multi-stage ``retrieve and re-rank'' framework as a novel solution to ICD indexing, via a hybrid discrete retrieval method, and re-rank retrieved candidates with contrastive learning that allows the model to make more accurate predictions from a simplified label space. The retrieval model is a hybrid of  auxiliary knowledge of the electronic health records (EHR) and  a discrete retrieval method (BM25), which efficiently collects high-quality candidates. In the last stage, we propose a label co-occurrence guided contrastive re-ranking model, which re-ranks the candidate labels by pulling together the clinical notes with positive ICD codes. 
%In the retrieval phase, auxiliary knowledge that includes different clinical code terminologies and drug prescriptions, and the discrete retrieval efficiently collect the high-quality candidates. In the re-ranker phase, contrastive learning constructs positive samples for clinical text under the guidance of the label co-occurrence. By pulling together the clinical notes with its corresponding positive sample, the text encoder is able to autonomously develop a text representation that is sensitive to label co-occurrence relationships.
Experimental results show the proposed method achieves state-of-the-art performance on a number of measures on the MIMIC-III benchmark. 
\end{abstract}

\section{Introduction}
%Structured clinical information, which adopts the form of coded clinical data and relies on controlled indexing vocabularies like ICD-10, constitutes a pivotal resource for the application of statistical analysis techniques to patient data 
Electronic health records\footnote{\urlstyle{same}\url{https://www.cms.gov/Medicare/E-Health/EHealthRecords}} (EHRs) contain a comprehensive repository of essential administrative and clinical data pertinent to a person's care within a specific healthcare provider setting. %This data resource includes demographics, historical notes, progress notes, laboratory findings, diagnoses, and prescribed medications, which are invaluable information for clinical decision-making. 
In order to conduct meaningful statistical analysis, these EHR data are annotated with structured codes in a classification system known as \textit{medical codes}. The International Classification of Diseases\footnote{\urlstyle{same}\url{https://www.who.int/standards/classifications/classification-of-diseases}} (ICD) is one of the most widely-used coding systems, and it provides a taxonomy of classes, each uniquely identified by a code assigned to an episode of patient care. %ICD codes, which are maintained by the World Health Organization\footnote{\urlstyle{same}\url{https://www.who.int}} (WHO), are one of the most widely-used coding systems in the world. There are two types of codes in the ICD coding system, namely procedure codes\footnote{\urlstyle{same}\url{https://en.wikipedia.org/wiki/Procedure_code}} (these are used to identify  specific surgical, medical, or diagnostic interventions) and diagnosis codes\footnote{\urlstyle{same}\url{https://en.wikipedia.org/wiki/Diagnosis_code}} (these are used to identify diseases, disorders and symptoms).

The task of medical coding associates ICD codes with EHR documents. The {\em status quo} of assigning medical codes is a manual process,  which is labour-intensive, time-consuming, and error-prone \cite{xie-xing-2018-neural}. 
%Incorrect code assignments result in various challenges, including the need for process reviews, financial losses, increased labour expenditures, and delays in the reimbursement process. 
To reduce coding errors and cost, the demand for automated medical coding has become imperative. %Automatic medical coding has become a considerable research subject, with some early work dating back to the 1990s \cite{Larkey1996CombiningCI} and continuing to the contemporary advancements in deep neural natural language processing (NLP) techniques. The trajectory of research in this domain shows the evolution of strategies aimed at enhancing the efficiency and accuracy of medical coding processes. 
Previous deep learning approaches regarded medical coding as an extreme multi-label text classification problem \cite{Shi2017TowardsAI, mullenbach-etal-2018-explainable,   Baumel2018MultiLabelCO, 10.1145/3357384.3357897,yuan-etal-2022-code}, where an encoder is typically employed to learn the representations of the clinical notes and a label-specific binary classifier is subsequently constructed on top of the encoder for label predictions. However, some remaining difficulties have still posed immense challenges. First, clinical documents are lengthy (containing on average 1596 words in the MIMIC-III dataset) and noisy (including terse abbreviations, symbols, and misspellings). Second, the label set is extremely large and complex; for instance, in the $10^{th}$ ICD edition, there are over 130,000 codes\footnote{\urlstyle{same}\url{https://www.cdc.gov/nchs/icd/icd10cm_pcs.htm}}. %70,000 procedure codes and over 69,000 diagnosis codes 
Third, the distribution of ICD codes is extremely long-tailed; while some ICD codes occur frequently, many others seldom appear, if at all, because of the rarity of the diseases. For instance, among the 942 unique 3-digit ICD codes in the MIMIC-III dataset \cite{Johnson2016MIMICIIIAF}, the ten most common codes account for 26\% of all code occurrences and the 437 least common codes account for only 1\% of occurrences \cite{Bai2019ImprovingMC}. 
\begin{figure}[t]
\includegraphics[width=\columnwidth]{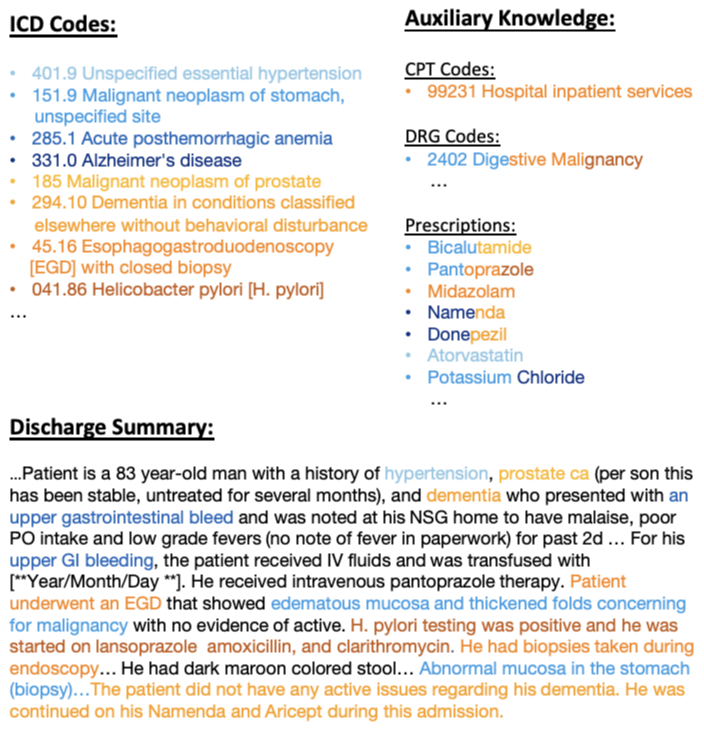}
\caption{An example of a medical record from the MIMIC-III dataset which includes the discharge summary, assigned ICD codes and auxiliary knowledge. We colour each code and its corresponding mentions in the discharge summary and auxiliary knowledge. We use the auxiliary knowledge of the notes to retrieve the candidate subset of the label space.}
\label{fig:1}
\end{figure}
To address the aforementioned challenges, we propose a novel multi-stage retrieve and re-rank framework, 
%retrieve-reduce-rerank framework, 
where the goal is to first generate a curated ICD list and then provide suggested ICD codes for a given medical record. In contrast to prior approaches, for instance, CAML~\cite{mullenbach-etal-2018-explainable}, MultiResCNN \cite{Li2020ICDCF} and KEPTLongformer \cite{yang-etal-2022-knowledge-injected}, that primarily consider ICD indexing as a multi-label text classification task, we introduce a new perspective that conceptualizes the task as a recommendation problem. More precisely, we first conduct a two-stage retrieval process leveraging auxiliary knowledge and BM25 %(first retrieve, then reduce) 
to obtain a small subset of candidate ICD codes from the large number of labels to alleviate issues caused by the label set and imbalanced label distribution. EHR auxiliary knowledge holds significant potential, but it has often been underutilized in prior studies. In addition to clinical texts, %an EHR document carries a wealth of auxiliary information, including various code systems beyond ICD codes and medication prescriptions. Specifically, 
our focus centers on two code terminologies: Diagnosis-Related Group codes\footnote{\urlstyle{same}\url{https://www.cms.gov/Medicare/Medicare-Fee-for-Service-Payment/AcuteInpatientPPS/MS-DRG-Classifications-and-Software}} (DRG) and Current Procedural Terminology codes\footnote{\urlstyle{same}\url{https://www.ama-assn.org/amaone/cpt-current-procedural-terminology}} (CPT), as well as patient prescribed medications. These external sources can serve as robust indicators for predicting ICD codes. For instance, within a drug prescription, the presence of a medication like ``Namenda'' can strongly imply a likelihood of Alzheimer's disease, as depicted in Figure \ref{fig:1}. Subsequently, we design a re-ranking model via co-occurrence guided contrastive learning to refine the candidate set, which can deal with lengthy clinical notes and generate semantically meaningful representations via the pre-trained language model and leverage code co-occurrence to generate co-occurrence-aware label representations. The co-occurrence of codes in clinical texts yields valuable insights into the interconnections among different diseases or conditions. As illustrated in Figure \ref{fig:1}, the code for ``Dementia in conditions classified elsewhere without behavioral disturbance'' (294.10) can be easily found in the text; however, inferring the code ``Alzheimer's disease'' (331.0) presents a more intricate challenge with less explicit clues. Fortunately, a robust association exists between these two diseases, with ``Alzheimer's disease'' serving as a prevalent cause of ``dementia''. This linkage can be effectively captured as these two diseases frequently co-occur within the clinical notes. This empowers us to gain a deeper understanding of the contexts, which could mitigate the limitation of long-tailed label distributions as rare labels might be suggested based on these relationships. We train the re-ranking model via contrastive learning as it has strong discriminative power that can extract features uniquely associated with each class, which empowers the model to make more accurate recommendations.  

To summarize, the major contributions of this paper are:
\begin{itemize}
\setlength{\itemsep}{0pt}
    \item We formalize the medical coding task as a recommendation problem and present a novel multi-stage retrieve and re-rank %retrieve-reduce-rerank 
    framework to make more accurate predictions by ruling out the irrelevant codes before ranking, rather than making direct predictions on the entire large label set. 
    \item To address the large label set and long-tailed distribution issues, in the two-stage retrieval process we use external knowledge and BM25 to retrieve a subset of candidate labels from the large label space. We further leverage the code co-occurrence in the re-ranking stage to capture the internal connections among the codes. 
    \item We apply contrastive learning in the re-ranking stage. It effectively pulls together the representations of a clinical note and its corresponding golden truth labels, which allows the model to make more accurate predictions. 
\end{itemize}
\section{Related Work}
%\subsection{Automatic ICD Indexing}
The automatic ICD indexing task is well established in the healthcare domain. %Early work in this area dates back to \citet{Larkey1996CombiningCI}, who proposed a method that fused three classifiers: K-nearest neighbour, relevance feedback, and a Bayesian independence classifier. This pioneering effort aimed to automatically assign ICD codes to dictated inpatient discharge summaries. Subsequently, \citet{10.1145/288627.288649} introduced a hierarchical model that hinged on the structure of the code topology, and calculated the cosine similarity between TF-IDF representations of the clinical texts and ICD codes. Over time, diverse strategies have been applied to the task of ICD coding, including rule-based methods \cite{10.5555/1572392.1572416,Farkas2008AutomaticCO} and statistical machine learning algorithms such as support vector machines \cite{lita-etal-2008-large}.
Extensive research using deep learning has been dedicated to  ICD indexing, including recurrent-based neural networks (RNNs), convolution-based neural networks (CNNs), and their variations \cite{mullenbach-etal-2018-explainable,Li2020ICDCF,Shi2017TowardsAI,xie-xing-2018-neural}. These 
%deep learning 
architectures 
%possess the remarkable capability to both 
are able to extract and categorize semantic features, reducing the need for medical domain expertise during the traditional feature selection stage seen in conventional algorithms \cite{9705116}. The ICD indexing task is formulated as a multi-label classification problem in these approaches. \citet{mullenbach-etal-2018-explainable} introduced a combination of CNN with an attention mechanism to effectively capture pertinent information within clinical texts for each ICD code. Building on this foundation, \citet{10.1145/3357384.3357897} enhanced the CNN attention model by integrating a multi-scale feature attention technique. Many CNN variants were subsequently introduced to address the challenges posed by lengthy and noisy clinical texts, including MultiResCNN \cite{Li2020ICDCF}, DCAN \cite{ji-etal-2020-dilated}, and EffectiveCAN \cite{liu-etal-2021-effective}. %MultiResCNN introduced a multi-filter residual CNN to capture text patterns of varying lengths, bolstered by a residual convolutional layer to expand the receptive field.  DCAN used a single filter and dilation operations to regulate the receptive field. EffectiveCAN combined a CNN-based encoder with squeeze-and-excitation networks and residual networks to extract the information across clinical texts. 
RNN-based models, renowned for their capacity to capture contextual information across input texts, have also been widely used for ICD indexing. \citet{Shi2017TowardsAI} proposed a character-aware Long Short-Term Memory (LSTM) recurrent network to learn the underlying representations of clinical texts. \citet{xie-xing-2018-neural} introduced a tree-of-sequences LSTM architecture alongside adversarial learning to capture hierarchical relationships among ICD codes. Additionally, \citet{Baumel2018MultiLabelCO} presented a Hierarchical Attention-Bidirectional Gated Recurrent Unit (HA-GRU) model, facilitating document labeling by identifying sentences relevant to each ICD code. LAAT \cite{ijcai2020-461} used a bidirectional Long-Short Term Memory (BiLSTM) encoder and a customized label-wise attention mechanism to cultivate label-specific vectors across distinct clinical text fragments. 

To address the hierarchical relationships intrinsic to ICD codes, Graph Convolutional Neural Networks (GCNNs) \cite{Kipf:2016tc} have emerged as a powerful tool. \citet{rios-kavuluru-2018-shot} and \citet{10.1145/3357384.3357897} used GCNNs to capture both the hierarchical interplay among ICD codes and the semantic information specific to each code. HyperCore \cite{Cao2020HyperCoreHA} took a comprehensive approach by considering both code hierarchy and code co-occurrence, employing GCNNs to learn code representations within the co-graph.

Incorporating external knowledge beyond ICD code information has also gained traction. \citet{Bai2019ImprovingMC} introduced a Knowledge Source Integration (KSI) model that integrates external knowledge from Wikipedia. This integration calculated matching scores between clinical notes and disease-related Wikipedia documents, in order to enrich the available information for ICD predictions. Additionally, \citet{yuan-etal-2022-code} proposed a Multiple Synonym Matching Network (MSMN) to use synonyms of ICD codes, enhancing the quality of code representation learning. Expanding on this, \citet{yang-etal-2022-knowledge-injected} integrated a pre-trained language model with three domain-specific knowledge sources: code hierarchy, synonyms, and abbreviations. This fusion of knowledge sources contributes significantly to the performance of ICD classification.

%\subsection{Retrieve and Rerank}
%The concept of retrieve-and-rerank is prevalent in recommendation systems, and generates recommendations in multiple stages, i.e., retrieval, ranking, and re-ranking. Each stage progressively refines the relevant items with a computationally more expensive but more precise model compared to the previous stages \cite{10.1145/3298689.3347000}. YouTube \cite{10.1145/2959100.2959190} used a two-stage information retrieval strategy, which first details a deep candidate generation model and then describes a separate deep ranking model. This framework has been widely adopted by later work \cite{Hron20212744}. The retrieval stage tends to use a discrete ranker (i.e., BM25) and  neural networks for re-ranking that mainly focuses on modeling the mutual influences between items \cite{10.1145/3477495.3532026}. The re-ranking method can be further divided into three categories: RNN-based \cite{10.1145/3209978.3209985}, Transformer-based \cite{10.1145/3397271.3401104}, and GNN-based methods \cite{10.1145/3340531.3412332}.  
\section{Method}
\subsection{A Multi-stage Framework}
\begin{figure*}[t]
\begin{center}
\includegraphics[width=\textwidth]{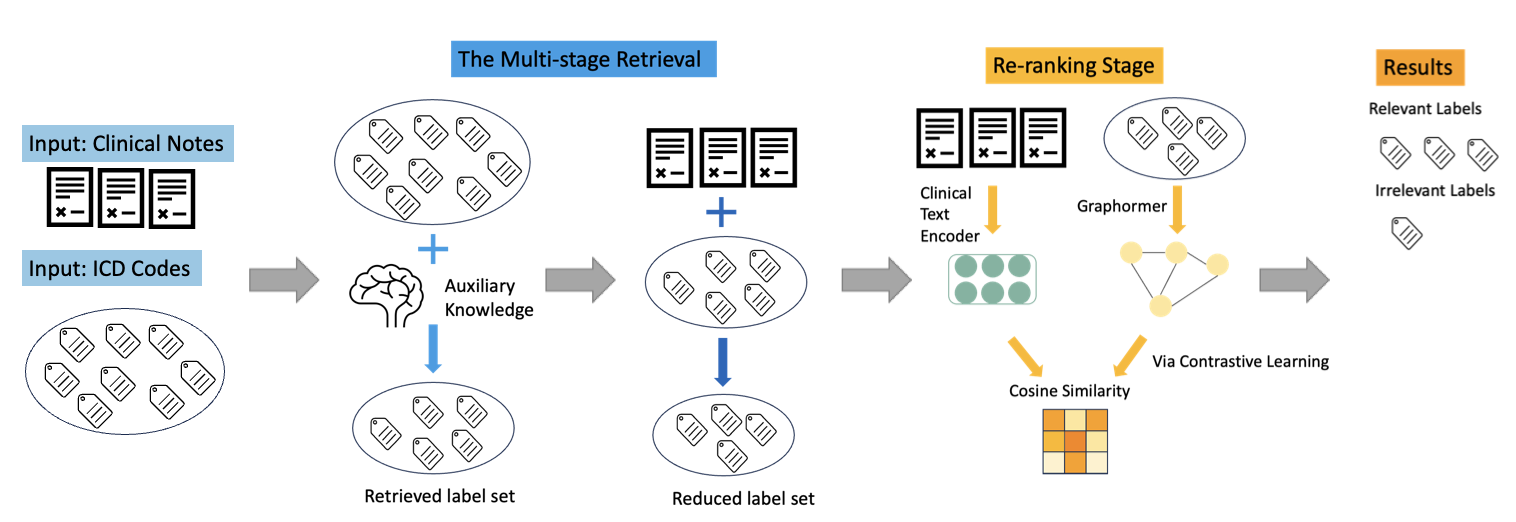}
\caption{Overview of the proposed multi-stage retrieve and re-rank %Retrieve-reduce-rerank 
framework. The model first leverages auxiliary knowledge and BM25 to retrieve a candidate list from the full label space, %then uses BM25 to reduce the labels in the candidate list, forming a more concise subset. Finally, 
then uses a re-rank model that leverages the code co-occurrence guided contrastive learning to generate the final relevant labels.}
\label{fig:2}
\end{center}
\end{figure*}
We formulate the medical coding task as a recommendation task given medical records $\mathcal{D} = \{d_1, d_2, ..., d_N \}$ and a set of ICD codes $\mathcal{Y} = \{y_1, y_2, ..., y_L\}$ with associated external auxiliary knowledge $\mathcal{K}$. We construct the label information as a graph structure $\mathcal{G}$, using  code co-occurrence relations, and we train a multi-stage recommender system $\mathcal{R}$, based on the text information $\mathcal{D}$, constructed label information $\mathcal{G}$, and the external auxiliary knowledge $\mathcal{K}$. The system $\mathcal{R}$ needs to predict the relevant labels given a document $d \notin \mathcal{D}$.
%The objective is to learn a predicted interaction function $\mathcal{R}: \mathcal{D} \rightarrow \mathcal{Y^r}$, where $\mathcal{Y^r}$ is a ranked label set.  
% the retrieval stage first generate a candidate list $\mathcal{C}_{init} \in \mathcal{Y}$, and then the re-ranking model aims to generate a re-ranking list, i.e., $\mathcal{C}_{final} \in \mathcal{C}_{init}$, that are most relevant to the medical records. 

In this section, we present a multi-stage retrieve and re-rank 
%novel retrieve-reduce-rerank 
framework for ICD indexing, which is shown in Figure \ref{fig:2}. Our model is composed of a two-stage retrieval process that uses auxiliary knowledge of the EHR and BM25
to obtain a shortened candidate list, %a reduction stage using BM25 that ranks the relevant ICD codes and selects top candidates as a more concise candidate list, 
and a re-ranking process that conducts code co-occurrence guided contrastive learning to further improve the recommended ICD list. 
\subsection{The Retrieval Stage} %Using Auxiliary Knowledge}

\paragraph{Using Auxiliary Knowledge}To retrieve the candidate list using auxiliary knowledge, we incorporate insights from three external sources of knowledge: diagnosis-related group (DRG) codes, current procedural terminology (CPT) codes, and medications prescribed to patients. DRG codes are used by hospitals and healthcare providers to classify patients into groups based on their diagnosis, treatment, and length of stay. These codes are used for reimbursement purposes, and they help determine the amount that healthcare providers are remunerated for their services. DRG codes are further classified into medical DRGs (which exclude operating room procedures) and surgical DRGs. CPT codes are used to describe medical procedures and services provided by healthcare providers. They provide a standardized way of documenting and billing for medical services. CPT codes are used by insurance companies to determine reimbursement rates for healthcare providers. Such code terminologies significantly contribute to the refinement of ICD indexing. Moreover, the medications prescribed to patients offer a wealth of predictive information for ICD codes. These prescriptions often mark the conclusion of a patient's care episode. As patients approach the conclusion of their treatment, the prescribed medications serve a critical role in managing their conditions. Consequently, these medications emerge as potent indicators of underlying health conditions or diagnoses. Their inclusion in the retrieval process greatly enhances the accuracy and relevance of the corresponding ICD code recommendations. The aforementioned auxiliary knowledge, such as DRG codes, CPT codes, and drug prescriptions, typically appears in the EHR data and is readily accessible. 

Given a clinical note $d$, we retrieve the candidate ICD list by calculating the auxiliary knowledge and label co-occurrence matrix using conditional probabilities, i.e., $P(y_{i}\,|\,k_{j})$, which denote the probabilities of occurrence of ICD $y_{i}$ when auxiliary knowledge $k_{j}$ appears. 
\begin{equation} \label{eq:6}
    P(y_{i}\,|\,k_{j}) = \frac{C_{y_{i}\cap k_{j}}}{C_{k_{j}}},
\end{equation}
where $C_{y_{i}\cap k_{j}}$ denotes the number of co-occurrences of $y_{i}$ and $k_{j}$, and $C_{k_{j}}$ is the number of occurrences of $k_{j}$ in the training set. To avoid the noise of rare co-occurrences, a threshold $\eta$ filters noisy correlations.
$\tilde{K}_{j}$ denotes the selected ICD set for auxiliary knowledge $j$. 
\begin{equation} \label{eq:7}
    \tilde{K}_{j} = \{y_{i} \vert P(y_{i} \vert k_{j}) > \eta, \;
    i = 1, ..., L\}, 
\end{equation}
where $L$ is the total number of ICD codes in the label set, and $\eta = 0.005$. We then join the ICD codes retrieved from the auxiliary knowledge co-occurrences for the DRG codes, CPT codes and prescribed drugs to form the candidate ICD subset $\mathcal{C}_{\mathrm{auxiliary}}$:
\begin{multline} \label{eq:10}
    %\small 
    \mathcal{C}_{\mathrm{auxiliary}} (d) = \\
    \tilde{K}_\mathrm{DRG} (d)\cup \tilde{K}_\mathrm{CPT} (d)\cup \tilde{K}_\mathrm{drug} (d),
\end{multline}
where $\mathcal{C}_{\mathrm{auxiliary}} \subseteq \mathcal{Y}$. 
%\subsection{The Relevance Reduction Stage Using BM25} 
\paragraph{Using BM25}The retrieval stage using auxiliary knowledge incorporates the co-relations between ICD codes and external knowledge, but ignores the relationship between clinical texts and labels. To increase the recall of the retrieval stage, we adopt BM25 \cite{10.5555/188490.188561} to allow lexical matching between the medical documents and labels on the retrieved candidate list $\mathcal{C}_{\mathrm{auxiliary}}$. Given a medical record $d$ and an ICD code $y$, the score between $d$ and $y$ is calculated as:
\begin{multline}
     %\small 
     \mathrm{BM25}(d, y) = \\ \sum_{w\in d \cap t_y} \mathrm{IDF}(w)\frac{\scriptstyle \mathrm{TF}(w, t_y) \cdot (k_1 + 1)}{\scriptstyle \mathrm{TF}(w, t_y) \cdot k_1(1-b+b\frac{|\mathcal{Y}|}{\textit{avgdl}})},
\end{multline}
and 
\begin{equation}
    \textit{avgdl} = \frac{1}{|\mathcal{Y}|}\sum_{y \in \mathcal{Y}}|t_y|,
\end{equation}
where $t_y$ represents the words in the label descriptors, $|\mathcal{Y}|$ is the length of the label descriptors in words, \textit{avgdl} is the average length of text information in the label. 

When the BM25 score between $d$ and $y_{i}$ exceeds a certain threshold $\theta$, we add $y_{i}$ as a candidate of $d$:
\begin{multline}
   %\small 
   \mathcal{C}_{\mathrm{BM25}} (d) = \\
   \{{y_{i}} \vert \mathrm{BM25}(d, y_{i}) > \theta, y_{i} \in \mathcal{C}_{\mathrm{auxiliary}}\}, 
\end{multline}
where $\theta = 200$. Given a clinical note $d$, its candidate ICD set is first generated by using the auxiliary knowledge in the retrieval stage and then reduced by using BM25, where $\mathcal{C}_{\mathrm{BM25}} \subseteq \mathcal{C}_{\mathrm{auxiliary}}$ and $\mathcal{C}_{\mathrm{auxiliary}} \subseteq \mathcal{Y}$. 
\subsection{The Re-ranking Stage} %Via Code Co-occurrence Guided Contrastive Learning}
%, build our re-ranker upon Clni pre-trained language models via a code co-occurrence guided contrastive learning. 

\paragraph{Clinical Text Encoder} Encouraged by the success of the pre-trained language model Longformer \cite{Beltagy2020Longformer} in  dealing with longer texts, we use Clinical-Longformer \cite{li2023comparative},  specifically pre-trained in the medical domain, as a text encoder. Given a medical document $d$ as input that consists of a sequence of tokens:
\begin{equation}
    d = \{[\texttt{CLS}], x_1, x_2, ..., x_{n-2}, [\texttt{SEP}]\},
\end{equation}
where $[\texttt{CLS}]$ and $[\texttt{SEP}]$ are two special tokens that indicate the beginning and end of the sequence, and $n$ is the sequence length, the Clinical-Longformer encodes the tokens and outputs the hidden representations for each token: 
\begin{equation}
    H_{\mathrm{hidden}} = \mathrm{ClinicalLongformer}(d),
\end{equation}
where $H_{\mathrm{hidden}} \in \mathbb{R}^{n \times h_e}$, and $h_e$ is the hidden size. Following previous work \cite{wang-etal-2022-incorporating, yang-etal-2022-knowledge-injected}, we use the hidden state of the $\mathrm{[CLS]}$ token to represent the document, which is the first token of $H_{\mathrm{hidden}}$, denoted as $H_{\mathrm{T}}$.

\paragraph{Label Encoder}
The occurrence of two ICD codes together in clinical texts frequently indicates a simultaneous presence or a causal connection between specific diseases. This implies that the codes representing these interconnected diseases often manifest together within clinical notes. 
We employ a Graphormer \cite{ying2021do} to incorporate the co-occurrence relationships among ICD codes. Unlike the original GNN, Graphormer models graphs using Transformer layers \cite{10.5555/3295222.3295349} with spatial encoding and edge encoding, which could effectively encode the structural information (i.e., code co-occurrence) of a graph into the model.
%To effectively capture this co-occurrence relationships among ICD codes within clinical texts, w
We create a directed code co-occurrence graph $\mathcal{G} = (\mathcal{Y}, \mathcal{E})$, where node set $\mathcal{Y}$ is the labels and edge set $\mathcal{E}$ denotes the co-occurrence relations. This graph is constructed using the code co-occurrence matrix, which has been used as the edge matrix for the graph. We create the code co-occurrence matrix by using the correlated relationship between labels based on conditional probabilities. This approach encapsulates the interdependence between various ICD codes in a quantifiable manner, offering valuable insights into the underlying connections among disease codes within the clinical texts. 
%by calculating the conditional probabilities of the co-occurrence relationships between ICD codes associated with the same medical documents.
To be more specific, we calculate the probability of occurrence of label $y_j$ when label $y_i$ appears as follows:
\begin{equation}
    P(y_j\,|\,y_i) = \frac{C_{y_{i}\cap y_{j}}}{C_{y_{i}}}
\end{equation}
where $C_{y_{j}\cap y_{i}}$ denotes the number of co-occurrences of $y_{i}$ and $y_{j}$, and $C_{y_{i}}$ is the number of occurrences of $y_{i}$ in the training set. To facilitate graph construction, we binarize the correlation probability $P(y_j\,|\,y_i)$. This entails converting the probability values into binary values which indicates whether a correlation exists (or not) between two labels. The operation can be written as:
\begin{equation}
    \mathcal{E}_{ij} = 
    \begin{cases}
        0, \text{if } P(y_j\,|\,y_i) < \lambda \\
        1, \text{if } P(y_j\,|\,y_i) \ge \lambda, 
    \end{cases}
\end{equation}
where $\mathcal{E}$ is the binary correlation matrix that is used to form the edge set, and $\lambda$ is the hyper-parameter threshold to filter the noise edges. In our experiment, $\lambda = 1$, which means that a edge is formed when the two labels in each pair always appear together. 

To encode the graph $\mathcal{G}$, we first generate the initial node features using the ICD full descriptors for each code $y$ via Clinical-Longformer:
\begin{equation}
    \begin{gathered}
        \mathrm{y} = \{[\texttt{CLS}], x_1, x_2, ..., x_{n-2}, [\texttt{SEP}]\}, \\
        H_v =  \mathrm{ClinicalLongformer}(\mathrm{y}),
    \end{gathered}
\end{equation}
where $\mathrm{y}$ represents a sequence of words in the label descriptors of label $y$, $H_v \in \mathbb{R}^{n \times h_e}$, and $h_e$ is the hidden size. We use the hidden state of the first token ($\mathrm{[CLS]}$) to represent the initial node feature denoted as $H_{\mathrm{node}}^i$ for the $i^{th}$ label.

With all initial node features stacked as a matrix $V = \{H_{\mathrm{node}}^1, H_{\mathrm{node}}^2, ..., H_{\mathrm{node}}^L\}$, where $V \in \mathbb{R}^{h_{e} \times L}$, a standard self-attention layer is then used for feature migration. To leverage the structural information, a novel spatial encoding method is used to modify the Query-Key product matrix $A^{\mathcal{G}}$ in the self-attention layer:
\begin{equation}
    A^{\mathcal{G}}_{\textit{ij}} = \frac{(H_{\mathrm{node}}^iW^{\mathcal{G}}_{Q})(H_{\mathrm{node}}^jW^{\mathcal{G}}_{K})^{\intercal}}{\sqrt{h_{e}}} + b_{\phi(y_i, y_j)},
\end{equation}
where $W^{\mathcal{G}}_{Q}$ and $W^{\mathcal{G}}_{K}$ are layer-specific weight matrices, and $\phi(y_i, y_j)$ is the spatial relation between $y_i$ and $y_j$ in graph $\mathcal{G}$, and the function $\phi(\cdot)$ is defined as the connectivity between the nodes in $\mathcal{G}$, which is the co-occurrence relation among labels. $b_{\phi(y_i, y_j)}$ is a learnable scalar indexed by $\phi(y_i, y_j)$, and shared across all layers. The attention score $A^{\mathcal{G}}_{\textit{ij}}$, then, has been used to aggregate the multi-head attention for the final output: 
%To form the self-attention:
%\begin{equation}
%    \mathrm{Attention} = \mathrm{softmax(A^{\mathcal{G}}})(v_{i}W^{\mathcal{G}}_{V}),
%\end{equation}
%where $W^{\mathcal{G}}_{V}$ is the layer-specific weight matrices. Then multi-head attention 
\begin{equation}
    h^{l+1} = \mathrm{MHA}(\mathrm{LN}(h^{l}))+h^{l}, 
\end{equation}
where $\mathrm{LN}$ denotes the layer normalization, $\mathrm{MHA}$ denotes the multi-head self-attention, $h^{l}$ and $h^{l+1} \in \mathbb{R}^{L \times h_e}$ indicate the node representation of the $l^{th}$ and $(l+1)^{th}$ layers. We use the last layer to represent the label feature denoted as $H_L$. For more details on the full structure of Graphormer, please refer to the original paper \cite{ying2021do}.   
\paragraph{Contrastive Learning for Re-ranking}
Now, we construct a code co-occurrence guided contrastive learning framework. Unlike supervised learning that aims to understand ``what is what'', contrastive learning adopts a different perspective by learning ``what is similar or dissimilar to what''. %In general, contrastive learning aims to pull together the positive samples in the embedding space and push apart the negative ones, which could effectively construct meaningful representations. By adopting contrastive learning, the re-ranking model has been enforced to generate closely aligned representations of the clinical notes and their corresponding ground truth labels within the embedding space. 
In our problem setting, we focus on the distances between a clinical document and its associated ICD codes, rather than solely between samples themselves. We consider the ground truth labels as positive samples, while the negative samples comprise all the other labels within the label space. Given $H_{\mathrm{T}}$, the representation for a clinical note $d$, and the set of representations of its corresponding ICD codes denoted as $H^{+}_{\mathrm{L}}$, %we denote all other ICD codes in the label space as $H^{-}_{\mathrm{L}}$. 
we denote the representations of $N$ negative ICD codes randomly chosen from the ICD codes of the documents in the batch (batch size is $N$), which are not ICD codes of document $d$, as $H^{-}_{\mathrm{L}}$.
Contrastive learning aims to learn the effective representations by pulling $d$ and $H^{+}_{\mathrm{L}}$ together while pushing apart $d$ and $H^{-}_{\mathrm{L}}$, represented as $S$ and $D$, respectively, in the equation below. The contrastive loss can be defined as:
\begin{equation}
    \mathcal{L} = -\mathrm{log} \frac{
    S / \tau}{S / \tau + D / \tau}, 
\end{equation}
where $S = \exp(\sum_{c \in L^+_{\mathrm{L}}} \cos(H_{\mathrm{T}}, c) / |H^{+}_{\mathrm{L}}|)$, $D = \exp(\sum_{c' \in L^-_{\mathrm{L}}} \cos(H_{\mathrm{T}}, c') / N)$, and $\tau$ is the temperature hyper-parameter. %, and $N$ is the number of clinical notes in a mini-batch. The re-ranking model is thus trained by minimizing this contrastive loss. 
During inference, a comparison is conducted by measuring the distance between the query clinical note and ICD codes in the embedding space, which ranks the ICD codes and then provides recommendations of the potential ICD candidates. 

\section{Experiments}
\subsection{Dataset and Pre-processing}
We conduct our experiments on the publicly available benchmark MIMIC-III \cite{Johnson2016MIMICIIIAF} dataset that contains a variety  of patient data types, including discharge summaries, demographic details, interventions, laboratory results, physiologic measures, and medication information. Following previous work, we are interested in the de-identified discharge summaries with annotated ICD-9 codes. 
There are 52,722 discharge summaries and 8,922 unique ICD-9 codes in the dataset. We mainly use three major data resources from the dataset: (1) de-identified discharge summaries (from the NOTEEVENTS table); (2) ICD-9 codes (from DIAGNOSES\_ICD and PROCEDURES\_ICD tables); and (3) auxiliary knowledge including DRG codes, CPT codes and drug prescriptions (from DRGCODES, CPTEVENTS, and PRESCRIPTIONS tables). 

To preprocess the clinical notes, we first remove all de-identified information, then replace punctuation and atypical alphanumerical character combinations (e.g., `3a', `4kg') with white space, and lowercase every token. We truncate the discharge summaries at a maximum length of 4000 tokens. We follow \citet{mullenbach-etal-2018-explainable} to form two settings: full codes (MIMIC-III-full) and top-50 frequent codes (MIMIC-III-top 50). In MIMIC-III-full, there are 47,719 discharge summaries for training, with 1,632 for validation, and with 3,372 for testing. 

\subsection{Implementation and Evaluation}
We implement our model in PyTorch \cite{NEURIPS2019_9015} on a single NVIDIA A100 40G GPU. We use the Adam optimizer and early stopping strategies using the Micro-F1 score over the validation set as the stopping criterion to avoid over-fitting. We set the initial learning rate as 5e-5 with batch size 16. We choose a learning rate scheduler which is warmed up with cosine decay, and the warm up ratio is set to 0.1. % We evaluate with 5 different random seeds for the model and report the average test results. 
Our code is available at \urlstyle{same}\url{https://github.com/xdwang0726/ICD-contrastive-curriculum}.

For evaluating the performance of our proposed model, we employ three commonly used metrics: F1-score (Micro and Macro), AUC (Micro and Macro), and precision at $K$ ($\mathrm{P@K}$). %The detailed computations of evaluation metrics can be found in the Appendix\ref{sec:appendix}.

\section{Results and Discussion}
\begin{table*}[t]
\resizebox{\textwidth}{!}{
\begin{tabular}{c cccccc  ccccc}
\hline
\multirow{3}{*}{Models} & \multicolumn{6}{c}{MIMIC-III-full}                                         & \multicolumn{5}{c}{MIMIC-III-top 50}                                    \\ \cline{2-12} 
                        & \multicolumn{2}{c}{AUC} & \multicolumn{2}{c}{F1} & \multicolumn{2}{c}{P@K} & \multicolumn{2}{c}{AUC} & \multicolumn{2}{c}{F1} & \multirow{2}{*}{P@5} \\ \cline{2-11}
                        & Macro      & Micro      & Macro      & Micro     & P@8        & P@15       & Macro      & Micro      & Macro      & Micro     &                      \\ \hline
CAML \cite{mullenbach-etal-2018-explainable}                   & 0.895      & 0.986      & 0.088      & 0.539     & 0.709      & 0.561      & 0.875      & 0.909      & 0.532      & 0.614     & 0.609                \\
DR-CAML \cite{mullenbach-etal-2018-explainable}                & 0.897      & 0.985      & 0.086      & 0.529     & 0.690      & 0.548      & 0.884      & 0.916      & 0.576      & 0.633     & 0.618                \\
MultiResCNN \cite{Li2020ICDCF}             & 0.910      & 0.986      & 0.085      & 0.552     & 0.734      & 0.584      & 0.899      & 0.928      & 0.606      & 0.670     & 0.641                \\
LAAT \cite{ijcai2020-461}                   & 0.919      & 0.988      & 0.099      & 0.575     & 0.738      & 0.591      & 0.925      & 0.946      & 0.666      & 0.715     & 0.675                \\
Joint-LAAT \cite{ijcai2020-461}             & 0.921      & 0.988      & 0.107      & 0.575     & 0.735      & 0.590      & 0.925      & 0.946      & 0.661      & 0.716     & 0.671                \\
EffectiveCAN  \cite{liu-etal-2021-effective}          & 0.915      & 0.988      & 0.106      & 0.589     & 0.758      & 0.606      & 0.915      & 0.938      & 0.644      & 0.702     & 0.656                \\
MSMN \cite{yuan-etal-2022-code}                   & \textbf{0.950}      & 0.992      & 0.103      & 0.584     & 0.752      & 0.599      & \textbf{0.928}      & 0.947      & 0.683      & 0.725     & 0.680                \\
KEPTLongformer \cite{yang-etal-2022-knowledge-injected}         & -          & -          & \textbf{0.118}      & 0.599     & 0.771      & 0.615      & 0.926      & 0.947      & \textbf{0.689}      & 0.728     & 0.672                \\ \hline
Ours                    &  0.949     & \textbf{0.995}      &   0.114    &   \textbf{0.603}   & \textbf{0.775}      & \textbf{0.623}    &    0.927        &    \textbf{0.947}        &    0.687       &    \textbf{0.732}     & \textbf{0.685} \\
\hline                   
\end{tabular}}
\caption{Comparison to previous methods across three main evaluation metrics MIMIC-III dataset. Bold: the optimal values.} \label{table:1}
\end{table*}

In order to asses the efficacy of our proposed framework, we compare with the existing state-of-the-art (SOTA) models, as outlined in Table \ref{table:1}. %Each row in the table corresponds to the evaluation metrics for a specific method. 
The top score for each metric is denoted in bold. As shown, our model outperforms in the majority of evaluation metrics, with the exception of Macro-AUC and Macro-F1 on MIMIC-III-full and MIMIC-III-top 50. Notably, our model achieves comparable performance on Micro-F1 and Micro-AUC, and improves precision at $K$ on both MIMIC-III-full and MIMIC-III-top 50. These results provide solid evidence to validate the efficacy of integrating auxiliary knowledge in the retrieval stage and leveraging code co-occurrence guided contrastive learning in the re-ranking stage. 

\begin{figure*}[t]
\begin{center}
\includegraphics[width=\textwidth]{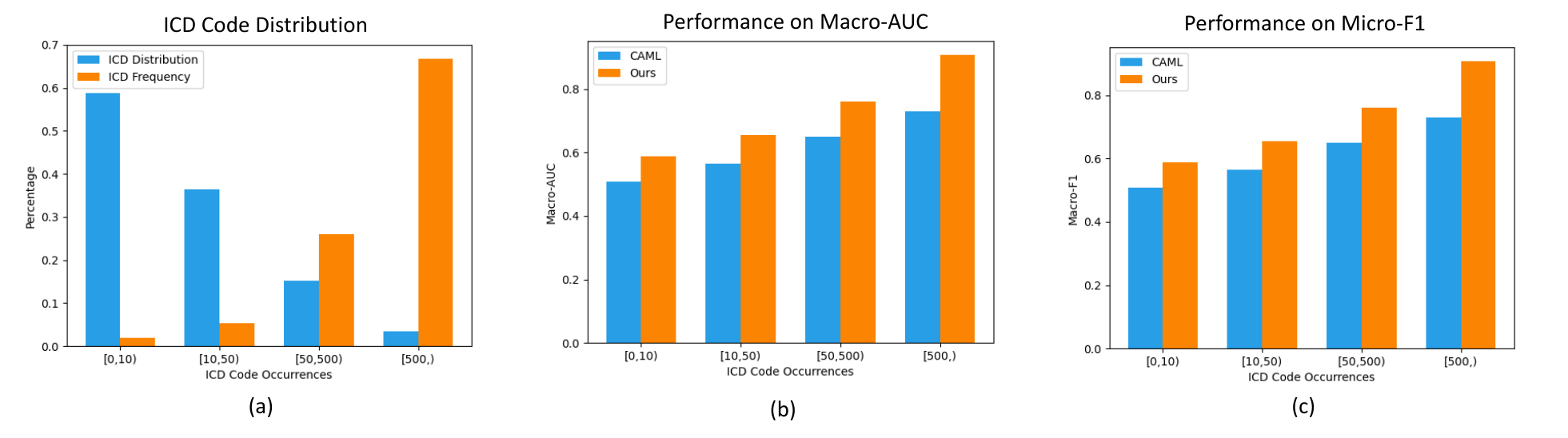}
\caption{(a) ICD code distribution. (b) Macro-AUC performance comparison of our model and CAML on ICD codes at different frequency. (c) Micro-F1 performance comparison of our model and CAML on ICD codes at different frequency.}
\label{fig:3}
\end{center}
\end{figure*}

As the occurrence frequencies of the ICD codes are imbalanced, our focus lies in assessing the efficacy of our model, specifically on the infrequently appearing ICD codes. We categorize the ICD codes into four groups based on their occurrences in the training set: [0, 10), [10, 50), [50, 500), and [500,). Figure \ref{fig:3} illustrates the distribution of ICD codes and their occurrence percentages across the four categorized groups in the training set, which show that the distribution of ICD frequency is highly biased, conforming to a long-tail distribution. Figures \ref{fig:3}b and \ref{fig:3}c present the performance of our model on MIMIC-III-full in comparison to the CAML baseline \cite{mullenbach-etal-2018-explainable} across the four ICD groups on $\text{Macro-AUC}$ and $\text{Micro-F1}$, respectively. Our model demonstrates significant improvements for both frequent and infrequent labels on both metrics.
%We aim to have a comprehensive investigation of different parts within our proposed model, and we are interested in gaining the insights into

To confirm the specific contributions of these modules in terms of enhancing both the effectiveness and robustness of the model, we conduct ablation studies 
%analyses that involve isolating certain parts from our model. Specifically, we conduct controlled experiments 
with three different settings: (a) we examine the effectiveness of using auxiliary knowledge in the retrieval stage by removing the retrieval stage and rank the ICD codes on the whole label set; (b) we examine the influence of different embedding methods by replacing the Clinical-Longformer with Clinical-BERT; and (c) we test the effectiveness of label embedding by replacing the encoding of the label with the average of word embeddings in the label descriptors. The experimental results are shown in Table \ref{table:3}. We also conduct case studies to qualitatively understand the effects of incorporating the label co-occurrence and the auxiliary knowledge. Two case studies have been presented in Appendix \ref{sec:appendix}.
\begin{table}[t]
\resizebox{\columnwidth}{!}{
\begin{tabular}{ccccc}
\hline
\multirow{2}{*}{Methods} & \multicolumn{2}{c}{F1} & \multicolumn{2}{c}{P@K} \\ \cline{2-5} 
                        & Macro      & Micro      & P@8        & P@15       \\ \hline
Full Model               &  \textbf{0.114} & \textbf{0.603}      & \textbf{0.775}      & \textbf{0.623}      \\ \hline
w/o auxiliary knowledge     &     0.097       &     0.579       &    0.748       &      0.587     \\
embedded w/ Clinical-BERT    &   0.083    &   0.548    &    0.711    &  0.546  \\
w/o Graphormer       &   0.102    &   0.583    &   0.753    &  0.591     \\
\hline
\end{tabular}}
\caption{Ablation experiment results on the MIMIC-III-full. Bold: the optimal values.} \label{table:3}
\end{table}
% \paragraph{Performance on Tail Labels}
\paragraph{Effectiveness of Using Auxiliary Knowledge for Retrieval} We employ three distinct types of auxiliary knowledge in the retrieval stage: DRG codes, CPT codes, and drug prescriptions. As shown in Table \ref{table:3}, removing auxiliary knowledge leads to a decline in performance,  indicating the pivotal role of the retrieval stage. This outcome further provides evidence that external knowledge effectively addresses the challenge presented by a large pool of potential ICD codes. Through integrating external knowledge, the retrieval stage attains the capability to refine the candidate list using the co-occurrence relationships between ICD codes and the auxiliary knowledge, thereby amplifying both the efficiency and accuracy of the re-ranking stage. The selection of an appropriate candidate list for a given medical record hinges upon a hyper-parameter, specifically the threshold $\eta$ governing the co-occurrence between auxiliary knowledge and ICD codes. The choice of $\eta$ determines the candidate numbers that implicitly affect the overall performance of the model. Setting $\eta = 0.005$, the candidate list guarantees inclusion of 99.22\% of the gold-standard ICD codes, resulting in an average of 1,460 codes in the subset. Notably, this accounts for approximately one-sixth of the complete code set. A further reduction using BM25 limits the candidate list to 1,299 on average. 

%\begin{table}[h]
%    \centering
%    \begin{tabular}{c c c c c c}
%    \hline
%         &  0.05 & 0.01 & 0.008 & 0.005 & 0.0025\\
%    \hline
%    $\mathrm{Percent}_{\mathrm{gold}}$ & 37.9\% & 82.3\% & 86.9\% & 93.4\% & 97.8\%\\
%    $\mathrm{Avg}_{\mathrm{retrieved}}$ & 225 & 921 & 1,086& 1,460 &2, 222 \\
%    \hline
%    \end{tabular}
%    \caption{Caption}
%    \label{table:4}
%\end{table} 

\paragraph{Comparison of Clinical-Longformer and Clinical-BERT} Increasing the maximum token limit is important in the context of clinical notes analysis as clinical texts are lengthy. Specially, in the MIMIC-III dataset, the average length of the discharge summaries is 1,596. Given this substantial token volume in the clinical notes, encoding a maximum number of tokens prior to downstream analysis becomes a pivotal requirement, which facilitates robust and meaningful subsequent analysis. To test the effectiveness of using longer sequences, we compare the model performance of Clinical-Longformer and a BERT-based pre-trained language model (i.e., Clinical-BERT) which can encode a maximum of 512 tokens. As shown in Table \ref{table:3}, Clinical-Longformer substantially outperforms Clinical-BERT, indicating the importance of the maximum token limit on language models in the automatic medical coding task. 
\paragraph{Effectiveness of Learning Label Features Using Code Co-occurrence} The graph structure has been shown to be effective in modeling code correlations and Graphormer efficiently learns code representations. The findings presented in Table \ref{table:3} highlight the affirmative impact of integrating code co-occurrence into label representations. By using Graphormer, the model effectively captures and exploits the intricate connections and interdependencies among the labels, thereby improving the overall performance. This indicates that incorporating code co-occurrence information with Graphormer empowers the model to gain insights from the collaborative behaviours of the labels, consequently facilitating a more holistic comprehension of the underlying label co-relations.  

\section{Conclusion}
In this paper, we regard the medical coding task as a recommendation problem and present a novel multi-stage retrieve and re-rank %retrieve-reduce-rerank 
framework. The primary objective of the proposed framework is twofold: to construct a curated list of ICD codes and, subsequently, to further refine the candidate list for a given medical record. Specifically, we first conduct a two-step retrieval process, incorporating auxiliary knowledge and the BM25 algorithm. This approach retrieves a concise subset of the candidate list, mitigating the challenges of a very large and imbalanced label distribution. We then use a re-ranking model to refine the previously obtained candidate list, employing code co-occurrence guided contrastive learning. Experimental results demonstrate that our proposed framework outperforms the previous SOTA, which suggests that it provides more precise and contextually grounded ICD recommendations for the given medical records. In the future, our proposed framework may be extended with more external knowledge such as the Unified Medical Language System (UMLS) and code synonymy. 

\section*{Limitations}
Our usage of auxiliary knowledge is limited to external knowledge that includes DRG codes, CPT codes, and drug prescriptions, only. Other knowledge including disease-symptom, disease-lab relations, Unified Medical Language System (UMLS), and others, could also be potentially useful for the auto ICD coding task. We also acknowledge that the auxiliary knowledge we used is labeled by human annotators, which may require some extra effort. We are not quite sure about the workload for annotating different code terminologies, but we believe linking different code terminologies is important. 

Our study is constrained by its evaluation limited to MIMIC-III-full and MIMIC-III-top 50 datasets, primarily concentrated on common disease. To comprehensively assess the model's performance on rare diseases, future work could benefit from a curated list of rare diseases validated by domain experts.

\section*{Ethics Statement}
We are using the publicly available clinical dataset MIMIC-III, which contains de-identified patient information. We do not see any ethics issue here in this paper.

\section*{Acknowledgements}
We would like to thank all reviewers for their comments, which helped improve this paper considerably. Computational resources used in preparing this research were provided, in part, by Compute Ontario\footnote{\urlstyle{same}\url{https://www.computeontario.ca}}, Digital Research Alliance of Canada\footnote{\urlstyle{same}\url{https://ccdb.alliancecan.ca}}, the Province of Ontario, the Government of Canada through CIFAR, and companies sponsoring the Vector Institute\footnote{\urlstyle{same}\url{https://www.vectorinstitute.ai/partners}}. This research is partially funded by The Natural Sciences and Engineering Research Council of Canada (NSERC) through a Discovery Grant to R. E. Mercer. F. Rudzicz is supported by a CIFAR Chair in AI.
\bibliography{anthology,custom}

\appendix

\section{Case Studies}
\label{sec:appendix}
\begin{figure}[ht!]
\includegraphics[width=\columnwidth]{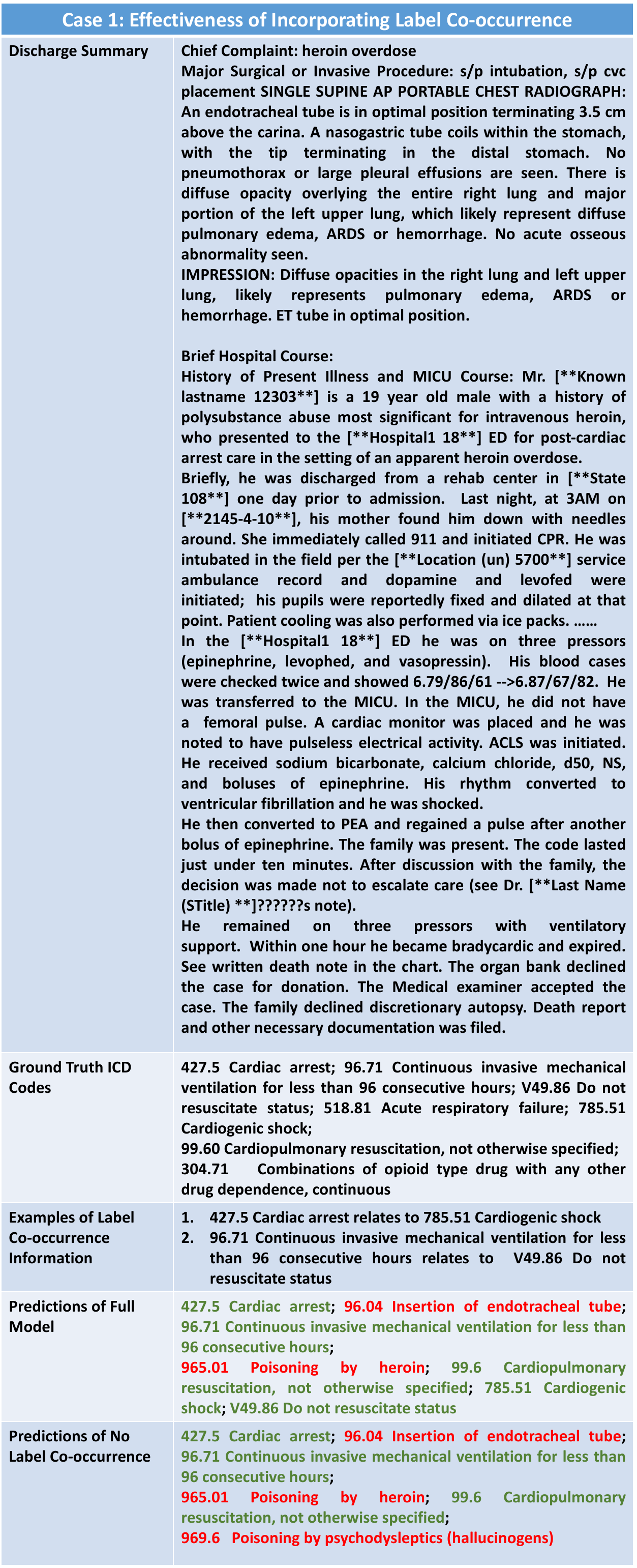}
\caption{Case study on the effectiveness of incorporating label co-occurrence. Correctly predicted labels are marked in green and the incorrect ones are marked in red.}
\label{fig:4}
\end{figure}
\begin{figure}[ht!]
\includegraphics[width=\columnwidth]{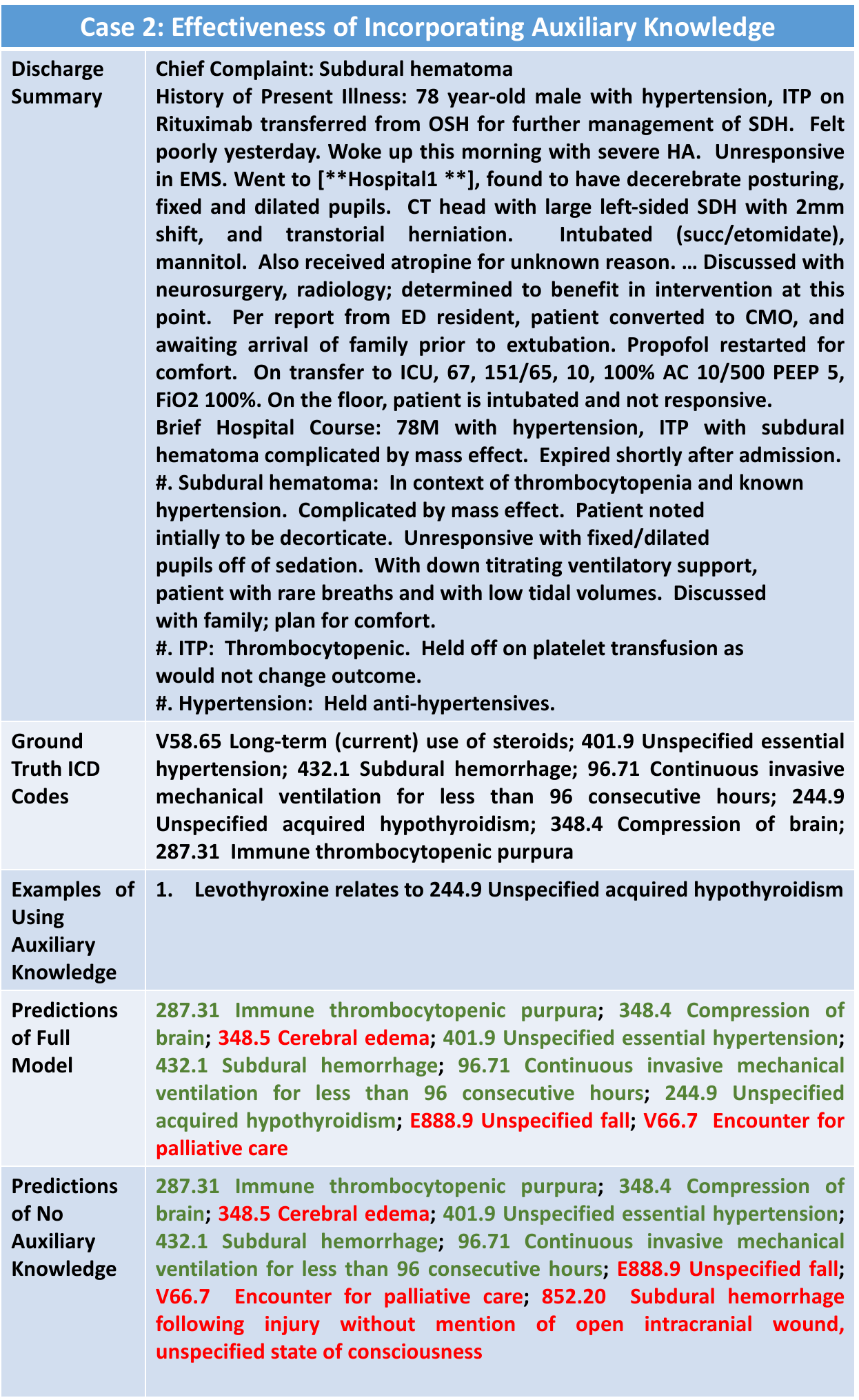}
\caption{Case study on the effectiveness of incorporating auxiliary knowledge. Correctly predicted labels are marked in green and the incorrect ones are marked in red.}
\label{fig:5}
\end{figure}

We conducted case studies to qualitatively explore the impacts of integrating label co-occurrence (illustrated in Figure \ref{fig:4}) and auxiliary knowledge (depicted in Figure \ref{fig:5}). We compared the full model with models that did not integrate the label co-occurence and the external knowledge on the predictions of two patient records. For each patient, we present the discharge summary, ground truth ICD codes, label co-occurrence information, and auxiliary knowledge information, along with the predicted ICD codes from the full model and ablated models.  

In Case 1, the ground truth ICD codes `785.51 Cardiogenic shock' and `V49.86 Do not resuscitate status' are not explicitly mentioned in the discharge summary. The observed label co-occurrence between `427.5 Cardiac arrest' and `785.51 Cardiogenic shock', as well as co-relation between `96.71 Continuous invasive mechanical ventilation for less than 96 consecutive hours' and `V49.86 Do not resuscitate status' provide strong indicators suggesting the presence of the codes `785.51' and `V49.86'. Without the label co-occurrence signals, the ablated model missed the predictions of codes `785.51' and `V49.86', indicating a failure to leverage latent label information.

In Case 2, the patient has been diagnosed with `244.9 Unspecified acquired hypothyroidism' with less explicit information in the discharge summary. Notably, the presence of the medication `Levothyroxine' in the drug prescription, an element of auxiliary knowledge, suggests that the patient is likely to have acquired hypothyroidism. The ablated model, lacking the auxiliary knowledge, misses the prediction of code  `244.9'. The aforementioned Cases 1 and 2 highlight the benefits of incorporating label co-occurrence and auxiliary knowledge, respectively.
\end{document}